\UseRawInputEncoding

\documentclass[letterpaper, 10 pt, conference]{ieeeconf}  

\IEEEoverridecommandlockouts                              

\usepackage{graphics} 
\usepackage{epsfig} 
\usepackage{mathptmx} 
\usepackage{times} 
\usepackage{booktabs}
\usepackage{tabularx}
\usepackage{graphicx}
\usepackage{cite}
\usepackage{amsmath,amssymb,amsfonts}
\usepackage{algorithmic}
\usepackage{algorithm}
\usepackage{textcomp}
\usepackage{xcolor}
\usepackage{soul}
\usepackage{multirow}

\title{\LARGE \bf
XR-DT: Extended Reality-Enhanced Digital Twin for Safe Motion Planning via Human-Aware Model Predictive Path Integral Control}

\author{
 	\parbox{\textwidth}{%
 		\centering
 		Tianyi Wang$^{1}$\textsuperscript{*}, Jiseop Byeon$^{1}$\textsuperscript{*}, Ahmad Yehia$^{1}$\textsuperscript{*}, Yiming Xu$^{2}$, Jihyung Park$^{2}$, \\ Tianyi Zeng$^{3}$, Sikai Chen$^{4}$, Ziran Wang$^{3}$, Junfeng Jiao$^{2}$, Christian Claudel$^{1\dag}$%
 	}%
 	\thanks{\textsuperscript{\dag}Corresponding author: Christian Claudel.}%
    \thanks{\textsuperscript{*}These authors contributed equally to this work.}%
    \thanks{$^{1}$Department of Civil, Architectural, and Environmental Engineering, The University of Texas at Austin, Austin, TX 78712, USA.
 		{\tt\small \{bonny.wang, jsbyeon, ahmad.yehia, christian.claudel\}@utexas.edu}}%
 	\thanks{$^{2}$School of Architecture, The University of Texas at Austin, Austin, TX 78712, USA.
 		{\tt\small \{yiming.xu, jihyung803\}@utexas.edu, jjiao@austin.utexas.edu}}%
    \thanks{$^{3}$School of Civil and Construction Engineering, Purdue University, West Lafayette, IN 47907, USA.
 		{\tt\small \{zeng366, ziran\}@purdue.edu}}%
    \thanks{$^{4}$Department of Civil and Environmental Engineering, University of Wisconsin-Madison, Madison, WI 53706, USA.
 		{\tt\small sikai.chen@wisc.edu}}%
 }

\begin{document}

\maketitle
\thispagestyle{empty}
\pagestyle{empty}

\begin{abstract}

As mobile robots increasingly operate alongside humans in shared workspaces, ensuring safe, efficient, and interpretable Human-Robot Interaction (HRI) has become a pressing challenge. 
While substantial progress has been devoted to human behavior prediction, limited attention has been paid to how humans perceive, interpret, and trust robots' inferences and how robots plan safe and efficient trajectories based on predicted human behaviors. 
To address these challenges, this paper presents XR-DT, an eXtended Reality-enhanced Digital Twin framework for mobile robots, which bridges physical and virtual spaces to enable bi-directional understanding between humans and robots. 
Our hierarchical XR-DT architecture integrates augmented-, virtual-, and mixed-reality layers, fusing real-time sensor data, simulated environments in the Unity game engine, and human feedback captured through wearable XR devices. 
Within this framework, we design a novel Human-Aware Model Predictive Path Integral (HA-MPPI) control model, an MPPI-based motion planner that incorporates ATLAS (Attention-based Trajectory Learning with Anticipatory Sensing), a multi-modal Transformer model designed for egocentric human trajectory prediction via XR headsets.
Extensive real-world experimental results demonstrate accurate human trajectory prediction, and safe and efficient robot navigation, validating the HA-MPPI's effectiveness within the XR-DT framework.
By embedding human behavior, environmental dynamics, and robot navigation into the XR-DT framework, our system enables interpretable, trustworthy, and adaptive HRI. 

\end{abstract}

\begin{figure}[ht]
  \centering
  \vspace{6pt}
  \includegraphics[width=\linewidth]{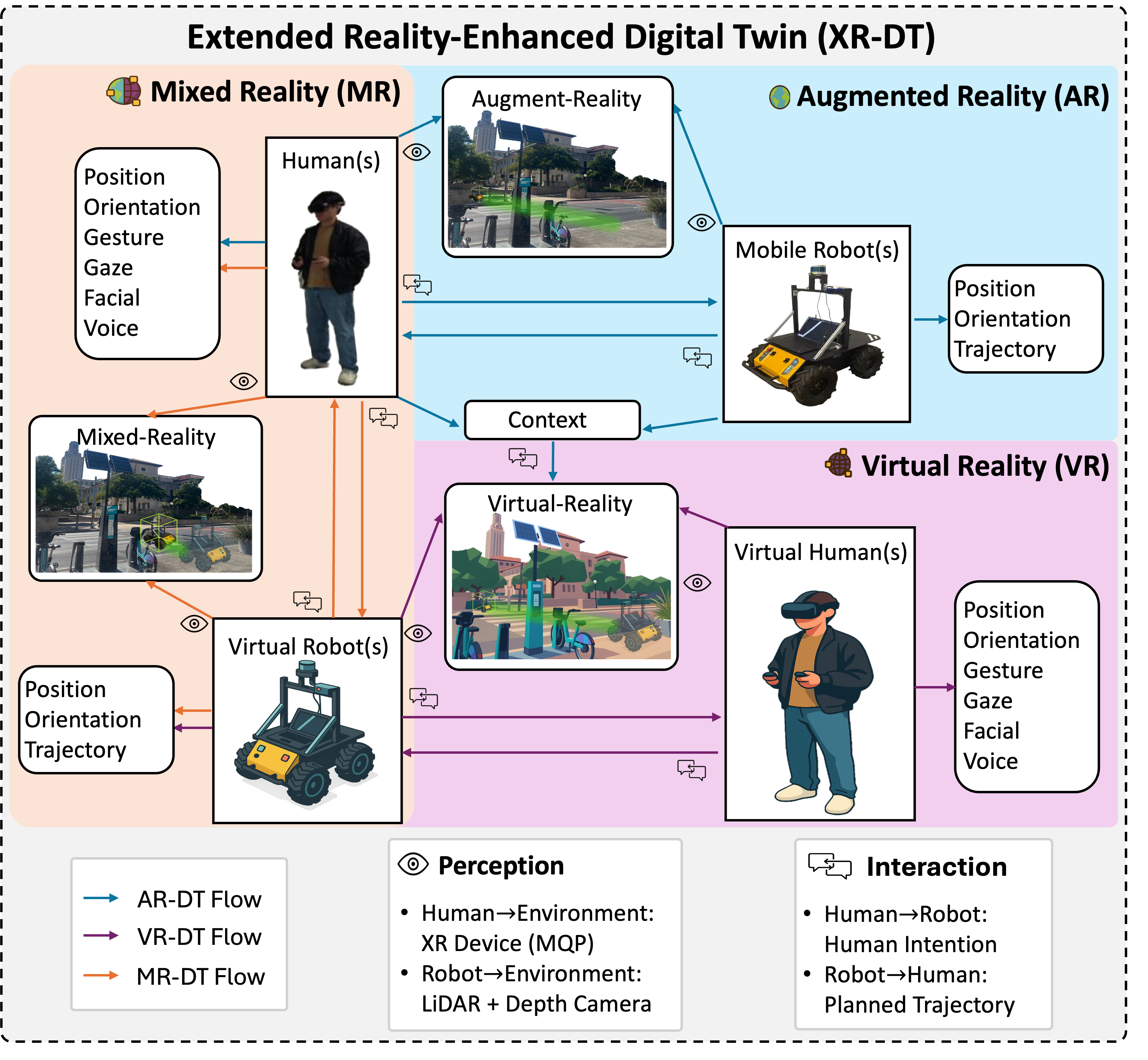}
  \caption{Extend reality-enhanced digital twin framework.}
  \label{Fig1}
\end{figure}

\section{Introduction}
Mobile robots, which share the working environment with humans, are expected to execute tasks efficiently, operate safely, and behave predictably to uncertain human behaviors~\cite{chen2009mixed}. 
Over the past decade, Human-Robot Interaction (HRI) has received increasing attention, with early work primarily emphasizing safety~\cite{ajoudani2018progress}. 
More recent applications require robots to communicate and collaborate with humans in shared workspaces, imposing challenging requirements on the robot's ability in understanding and predicting human intentions~\cite{luo2018unsupervised}. 
Many studies have shown the capability of robots to interpret humans' activities, while humans' ability to understand robots' internal inferences is often neglected~\cite{ravichandar2016human}. 
This gap limits the deployment of autonomous robot systems in safety-critical or socially embedded settings.

Within the vast paradigm of Cyber-Physical Systems (CPS), Digital Twin (DT) technology, which serves as a panoptic reflection of a physical body in the digital world, has been increasingly adopted to support robotic industries~\cite{perno2022implementation}. 
The benefits of using DTs include the reduced human workload and cost, the flexibility to test different algorithmic/architectural designs in various experimental scenarios, and the possibility to reuse the knowledge generated by DTs in future tasks.
However, current research mostly considered DT in pure Virtual Reality (VR), which made it impossible to realize real-time monitoring and synchronization of real-world activities, contrary to its original definition~\cite{serrano2023digital}.
Recent advances in wearable Augmented Reality (AR) technology have not only stimulated the development of DT, but also led to a trend of developing mutual communication methods to improve HRI~\cite{prattico2021comparing}.
For example, Project Aria~\cite{engel2023project} captured human proprioception data and advanced multi-modal egocentric perception research.
Theoretically, based on AR devices, a Mixed Reality (MR)-enhanced DT framework can simulate a comprehensive environment and realistic feedback from the robot, and then send it back to humans in the physical world. 
However, existing studies are limited to understanding specific HRI tasks with limited objectives, e.g., only considering hand-object interactions in manipulation and grasping, in a fixed indoor environment monitored in the training phase~\cite{kareer2025egomimic}. 
Although MR has shown great potential, such applications may seem impractical, as they still raise concerns about how non-expert users understand, trust, and act upon robot-generated insights.
Another essential aspect for HRI is the safe and efficient robot navigation in uncertain dynamic environments, which is challenging in predicting the future states of humans.
Optimization-based methods, particularly Model Predictive Control (MPC), are a key technology for human-aware motion planning, allowing robots to anticipate future events and adjust paths while considering multiple constraints, and multi-modal human motion prediction \cite{heuer2023proactive}.
There are two main MPC frameworks that can handle uncertainty in dynamic environments: (1) robust MPC \cite{bemporad2007robust}, which enforces safety guarantees by considering worst-case scenarios within bounded uncertainty sets; (2) stochastic MPC \cite{mesbah2016stochastic}, which employs chance constraints that probabilistically bound the likelihood of constraint violations within a specified confidence level.
However, robust MPC often falls into the ``frozen robot" problem due to its overly cautious behavior, while stochastic MPC allows for more flexible decision-making while maintaining a controlled level of risk. 
However, existing human-aware robot navigation methods often simplify assumptions--modeling uncertainty with Gaussian distributions \cite{zhu2019chance} or restricting the human dynamics to linear systems \cite{kruse2013human}--resulting in limited flexibility and generalization. 

To overcome the above problems, this paper presents \textbf{XR-DT} (Figure~\ref{Fig1}), an eXtend Reality (XR)-enhanced DT framework for mobile robots, to improve task performance by providing better awareness of the robot’s inference.
Besides, this paper proposes a novel Human-Aware Model Predictive Path Integral (\textbf{HA-MPPI}) control model (Algorithm~\ref{alg:ha_mppi}), to ensure safe and efficient robot navigation in uncertain dynamic environments.
The main contributions are as follows:
\begin{itemize}
\item{\textbf{Extend Reality-Enhanced Digital Twin Architecture}}: We utilize sensor-based knowledge in the real world to generate real-time human, robot, and environment in VR-enhanced DT; we provide a combination of real and virtual hints in AR-enhanced DT and MR-enhanced DT, respectively; and we allow humans to use wearable XR devices to send commands to robots, and the predicted robotic motion would be shown in XR-enhanced DT.
\item{\textbf{Human-Aware Model Predictive Path Integral Control Framework}}: We design a multi-modal human motion prediction model based on real-time 6 Degree-of-Freedom (DoF) pose, eye gaze, and egocentric RGB video collected by a wearable XR device; and then we integrate it into the MPPI-based motion planning algorithm and conduct navigation in heterogeneous space.
\end{itemize}

\section{Related Work}

\subsection{Mixed Reality-Enhanced Digital Twin}
MR was first proposed by Milgram and Kishino~\cite{milgram1994taxonomy}, which serves as the merging of physical and virtual worlds, allowing both to interact with each other. 
From the viewpoint of HRI, MR provides a new interface, providing seamless spatial interaction between humans and robots. 
Piumatti et al.~\cite{piumatti2017spatial} applied MR technology to improve the interactive experience of their HRI game. 
Wu et al.~\cite{wu2020mixed} proposed an MR-based control system for mobile robot path planning, supporting obstacle avoidance by adding interaction between the mobile robot and virtual objects. 
At the same time, mobile robots need to communicate their inner state, perceived world state, and planning of actions ahead of time~\cite{chakraborti2018projection}. 
Although many new MR technologies have emerged, most MR applications are still not as efficient as envisioned when dealing with complex tasks. 
MR aims to map the real-world environment into the virtual DT, but most of the current applications cannot achieve this. 

\subsection{Human-Aware Collision Avoidance}
Traditional MPC approaches achieve human-aware collision avoidance by considering online human motion detection \cite{chen2023trajectory}, incorporating safety-critical constraints into the optimization process \cite{mohamed2025chance}, and integrating human motion prediction into the MPC optimization objective \cite{akhtyamov2025social}.
Recent studies have also exploited multi-modal human motion predictions for generating robot motion \cite{chen2021reactive}.
Building upon Human-MPC \cite{hielscher2024towards}, a fast-embedded optimization method with 2D human motion prediction, context-aware MPC \cite{stefanini2024efficient} considered human activities and 3D human body poses.
MPPI \cite{williams2017model}, as a type of stochastic MPC method, stands out among alternative MPC methods as a promising control strategy for complex robotics systems with stochastic dynamics and uncertainty, benefiting from parallel sampling and the computational capabilities of GPUs to achieve optimized and real-time performance.
Risk-aware MPPI \cite{yin2023risk} was investigated for motion planning with probabilistic estimation uncertainty in partially known environments.
More recently, dynamic risk-aware MPPI \cite{trevisan2025dynamic} used Monte-Carlo sampling to approximate the chance constraints for a mobile robot navigating among dynamic agents.
However, a significant shortcoming in the field is the limited consideration of multi-modal human data in MPC despite the benefits they offer in terms of enhancing HRI and safety in robotic navigation.

\begin{figure*}[htbp!]
  \centering
  \includegraphics[width=\linewidth]{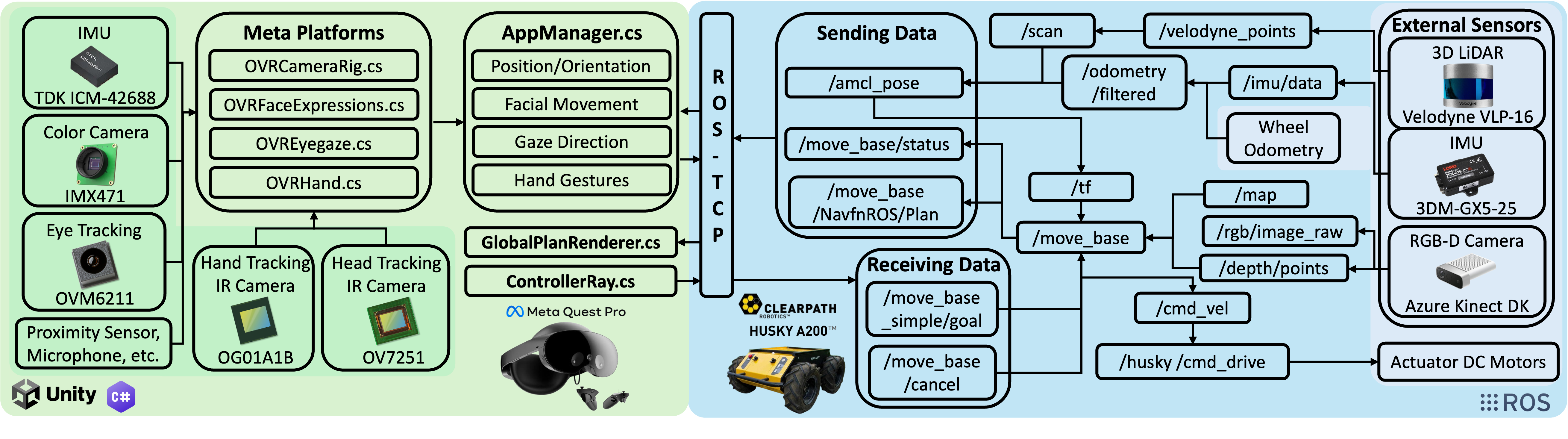}
  \caption{Overview of the integrated Unity-ROS operating system.}
  \label{Fig2}
\end{figure*}

\section{Preliminary}

\subsection{Problem Formulation}
\label{subsec:problem}

We consider a controlled mobile robot navigating in a planar workspace $\mathcal{W} \subseteq \mathbb{R}^2$ with nonlinear discrete-time dynamics given by:
\begin{equation}
    \boldsymbol{s}_{t+1} = f(\boldsymbol{s}_t, \boldsymbol{u}_t + \delta\boldsymbol{u}_t),
\end{equation}
where $f(\cdot)$ represents a nonlinear state-transition function, $\boldsymbol{s}_t = \left[\boldsymbol{p}_t, \theta_t\right] \in \mathbb{R}^{n_x}$ is the robot state at time step $t$, which contains its 2D position $\boldsymbol{p}_t = [x_t,y_t]^\top \in \mathbb{R}^2$ and orientation $\theta_t$, $\boldsymbol{u}_t \in \mathbb{R}^{n_u}$ denotes the control input of the robot at time step $t$, $\delta\boldsymbol{u}_t \sim \mathcal{N}(0,\Sigma_u)$ is a zero-mean Gaussian noise with covariance $\boldsymbol{\Sigma}_u \in \mathbb{R}^{n_u \times n_u}$, and scalars $n_x$ and $n_u$ are the number of state and input spaces of the robot, respectively. 

Given that each pedestrian is equipped with a wearable XR headset that provides real-time 6-DoF pose measurements. 
The human state at time step $t$ is represented as: 
\begin{equation}
h_t= \left[x_t,y_t,z_t,\phi_t,\theta_t,\psi_t\right]^\top, 
\label{eq:2}
\end{equation}
where $\left[x_t, y_t, z_t\right]^\top \in \mathbb{R}^3$ denotes the 3D position, and $\left[\phi_t,\theta_t,\psi_t\right]^\top \in \mathbb{R}^3$ corresponds to yaw, pitch, and roll angles, respectively. 
We observe the history of human states $ \mathcal{H} = \{h_{t-T_{obs}+1}, ..., h_{t}\}$ with $T_{obs} = 20$ frames and predict the future human states $\hat{\mathcal{H}} = \{\hat{h}_{t+1}, ..., \hat{h}_{t+Tpred}\}$ for $T_{pred} = 40$ frames.

The mobile robot navigates in an environment occupied by $N_h$ pedestrians, whose future trajectories are predicted by the multi-modal human motion prediction model presented in Section \ref{sec:atlas}. 
Safe motion planning in a dynamic and uncertain environment can be formulated as a chance-constraint collision avoidance problem \cite{trevisan2025dynamic}.
The objective of the stochastic MPC problem is to find an optimal control sequence $\boldsymbol{u_t} \in \mathbb{U}$ that generates a collision-free trajectory, guiding the robot from its initial state $\boldsymbol{s}_0$ to the desired goal state $\boldsymbol{s}_f$ while minimizing the cumulative cost function $J$ satisfying state, input, and probabilistic collision safety constraints.
The resulting finite-horizon (length $N$) optimization problem is given by:
\begin{subequations}
\begin{alignat}{2}
\min_{\boldsymbol{u} \in \mathbb{U}} \quad & J = \mathbb{E}\left[ \sum_{t=0}^{N-1}J_t(\boldsymbol{s}_t, \boldsymbol{u}_t, \delta\boldsymbol{u}_t) + J_T(\boldsymbol{s}_N)\right], \label{eq:3a} \\
\textrm{s.t.} \quad & \boldsymbol{s}_0 = \boldsymbol{s}_{\text{init}}, \quad \boldsymbol{s_t} \in \mathbb{S}, \boldsymbol{u_t} \in \mathbb{U}, \forall t \in \{0, 1, ..., N-1 \}, \\
 & \boldsymbol{s}_{t+1} = f(\boldsymbol{s}_t, \boldsymbol{u}_t + \delta\boldsymbol{u}_t), \quad \delta\boldsymbol{u}_t \sim \mathcal{N}(0,\boldsymbol{\Sigma}_u), \\
  &\mathbb{P} \left[ ||\boldsymbol{s}_t - \hat{\boldsymbol{h}}_t^o||_2 - r \geq 0\right] \geq 1-\sigma, \quad \forall o \in \{1,2, ..., N_h\},
\end{alignat}
\end{subequations}
where $J_t(\cdot)$ represents the stage cost of the robot, $J_T(\cdot)$ denotes the terminal cost of the robot, states $\boldsymbol{s}_t$ and inputs $\boldsymbol{u}_t$ are bounded by the state and input constraint sets $\mathbb{S}$ and $\mathbb{U}$, respectively, $\hat{\boldsymbol{h}}_t^o$ denotes the predicted position of the $o$-th pedestrian at time $t$, scalar $\sigma \in (0,1)$ is the desired risk level, and safety radius $r$ is the sum of the robot and human radii.

\subsection{Model Predictive Path Integral Control}

MPPI control is a Monte Carlo sampling-based, derivative-free optimization method for solving finite-horizon stochastic optimal control problems with nonlinear dynamics and non-convex cost functions \cite{williams2017model}.
In this work, MPPI is specialized to handle the chance-constrained collision avoidance problem formulated in Section \ref{subsec:problem}.
At each control cycle, a nominal control sequence over a finite horizon of length $N$ is maintained $\mathbb{U} = \{\boldsymbol{u}_0, \boldsymbol{u}_1, ..., \boldsymbol{u}_{N-1}\}$.
To generate potential system trajectories, a set of $K$ perturbed control sequences is generated by injecting zero-mean Gaussian noise into the nominal controls. 
For each sample $k$ and time step $t$, the sampled control is $\boldsymbol{u}_t + \delta\boldsymbol{u}_t^k$, and satisfies $\delta\boldsymbol{u}_t^k \sim \mathcal{N}(0,\Sigma_u)$.
Each sampled control sequence $k$ is forward-simulated through the stochastic system dynamics to generate a corresponding state trajectory $ \boldsymbol{\tau}_k = \{\boldsymbol{s}_1^k, \boldsymbol{s}_2^k, ..., \boldsymbol{s}_N^k\}$, for $k \in \{1,2,...,K\}$.
For each sampled trajectory $\boldsymbol{\tau}_k$, a \textit{cost-to-go} is evaluated using the same stage and terminal cost functions defined in the stochastic MPC formulation (Equation \ref{eq:3a}). 
The total cost associated with the $k$-th trajectory is:
\begin{equation}
J(\boldsymbol{\tau}_k) = \mathbb{E}\left[ \sum_{t=0}^{N-1}J_t(\boldsymbol{s}_t^k, \boldsymbol{u}_t, \delta\boldsymbol{u}_t^k) + J_T(\boldsymbol{s}_N^k)\right].
\end{equation}

MPPI then employs an importance sampling scheme to approximate the optimal control sequence. 
Each trajectory is assigned a weight $w_k$ according to an exponential transformation of its cost:
\begin{equation}
w_k = \exp \left(-\frac{1}{\lambda}\left[J(\boldsymbol{\tau}_k) - \min_{k \in K}J(\boldsymbol{\tau}_k)\right]\right),
\end{equation}
where $\lambda > 0$ is the inverse temperature parameter, which controls the selectivity of the weighted average of trajectories.
The final control input sequence $\boldsymbol{u}_t$ is then computed as a weighted sum of the sampled control sequences:
\begin{equation}
\boldsymbol{u}_t \leftarrow \boldsymbol{u}_t + \frac{\sum_{k=1}^K w_k \delta \boldsymbol{u}_t^k}{\sum_{k=1}^K w_k}.
\end{equation}

The resulting sequence is then smoothed using a Savitzky-Golay filter \cite{savitzky1964smoothing}, followed by applying the first control input $\boldsymbol{u}_0$ to the system, while the remaining sequence is shifted forward in time and used as a warm-start for the next optimization cycle.

\section{Methodology}

\subsection{Extended Reality-Enhanced Digital Twin}
\label{sec:system}

To enable real-time environmental understanding, intent sharing, and adaptive HRI, we design an XR-DT framework. 
As illustrated in Figure~\ref{Fig1}, the XR-DT framework consists of three workflows: AR-DT, VR-DT, and MR-DT Flow. 
The AR-DT and VR-DT each establish an independent perception-interaction loop, while the MR-DT integrates these two loops into a unified cross-world loop that enables bi-directional interaction between humans operating in the real world and robots planning in the virtual world. 

\subsubsection{Augmented Reality-Enhanced Digital Twin}

The AR-DT serves as the system's primary interface to the physical world, embedding both human and robot information directly into the real environment. 
Through a head-mounted XR device, the system captures multi-modal human data, while the mobile robot simultaneously provides its own sensor-based data, structured in a representation consistent with the VR-DT to ensure coherent cross-world state encoding.
These heterogeneous data streams are processed to infer both human and robot intentions, as well as the environmental context.
Based on these intermediate representations, the system estimates the future trajectories of both humans and robots. 
The AR-DT then overlays DT-derived cues, including robots' and humans' trajectories and semantic annotations, directly onto the user's egocentric view. 
This visualization transforms the physical environment into an interpretable and context-aware information space, enabling humans to anticipate robot behavior and allowing robots to infer human intent for safe and efficient motion planning.

\subsubsection{Virtual Reality-Enhanced Digital Twin}

The VR-DT functions as the simulation and predictive reasoning space by constructing a virtual representation of humans, robots, and their surrounding environment. 
This virtual world is initialized from a pre-built 3D DT model based on the human, robot, and environmental information provided by the AR-DT.
To maintain consistency with the physical world, the VR-DT is continuously updated using the robot's LiDAR-derived 3D spatial information, ensuring that virtual representations remain synchronized with essential environmental changes. 
At the core of the VR-DT, the system synthesizes multi-modal observations to model the virtual environment, infer human intention, and evaluate a diverse set of candidate robot behaviors.
By comparing the predicted outcomes across multiple hypothetical scenarios, humans can select an appropriate navigation strategy and task plan for the robot. 
This simulation environment provides a risk-free space for prediction, assessment, and behavioral optimization, enabling the robot to exhibit coherent, anticipatory, and safety-aware behavior before any action is executed in the physical world.

\subsubsection{Mixed Reality-Enhanced Digital Twin}

The MR-DT serves as the integrative layer of the XR-DT system, merging the independent perception-interaction loops of the AR-DT and VR-DT into a unified cross-world interaction space.
Specifically, it incorporates (1) long-horizon, simulation-driven predictions obtained through extensive scenario exploration from the VR-DT, and (2) context-rich, real-time information reflecting current human behaviors, environmental dynamics, and situational cues from the AR-DT. 
By jointly evaluating these two sources, the MR-DT synthesizes a coherent and context-aware system decision that is both feasible in the physical world and optimized with respect to predicted future outcomes. 
This decision is then projected back to the XR device interface, spatially aligned with the user's physical surroundings and visualized together with semantic annotations, predicted system responses, and interpretable reasoning cues. 
Through this unified representation, the MR-DT enables adaptive, interpretable, and collaborative human-robot operation with shared situational awareness in dynamic and uncertain environments.

\subsection{Operating System Overview}

An overview of the hardware and software components of the XR-DT system is provided in Figure~\ref{Fig2}.
The platform integrates XR devices and mobile robots, both of which communicate with a centralized Python server via Transmission Control Protocol/Internet Protocol (TCP/IP) connections. 

\subsubsection{Hardware}

The hardware system consists of a Meta Quest Pro (MQP) headset for capturing multi-modal human data and a Clearpath Husky A200 mobile robot equipped with wheel odometry, an IMU, an RGB-D camera, and a 3D LiDAR for environmental perception. 
Both platforms stream sensor data to a centralized Python server (Intel Core Ultra 9 285H, 32 GB RAM, NVIDIA Pro 2000 Blackwell) over TCP/IP for real-time perception fusion and HRI inference.

\subsubsection{Software}

The XR-DT processing pipeline is developed in the Unity game engine using the Meta XR SDK, which provides access to the headset's motion, gaze, hand, and facial tracking modules. 
These features are encoded and transmitted to the server using custom communication scripts. 
The robot operates under a ROS architecture, where each onboard sensor publishes synchronized messages that are forwarded to the server via dedicated TCP nodes. 

\subsubsection{Data Synchronization and Preprocessing}
\label{subsubsec:data}

\begin{figure}[htbp!]
  \centering
  \includegraphics[width=\linewidth]{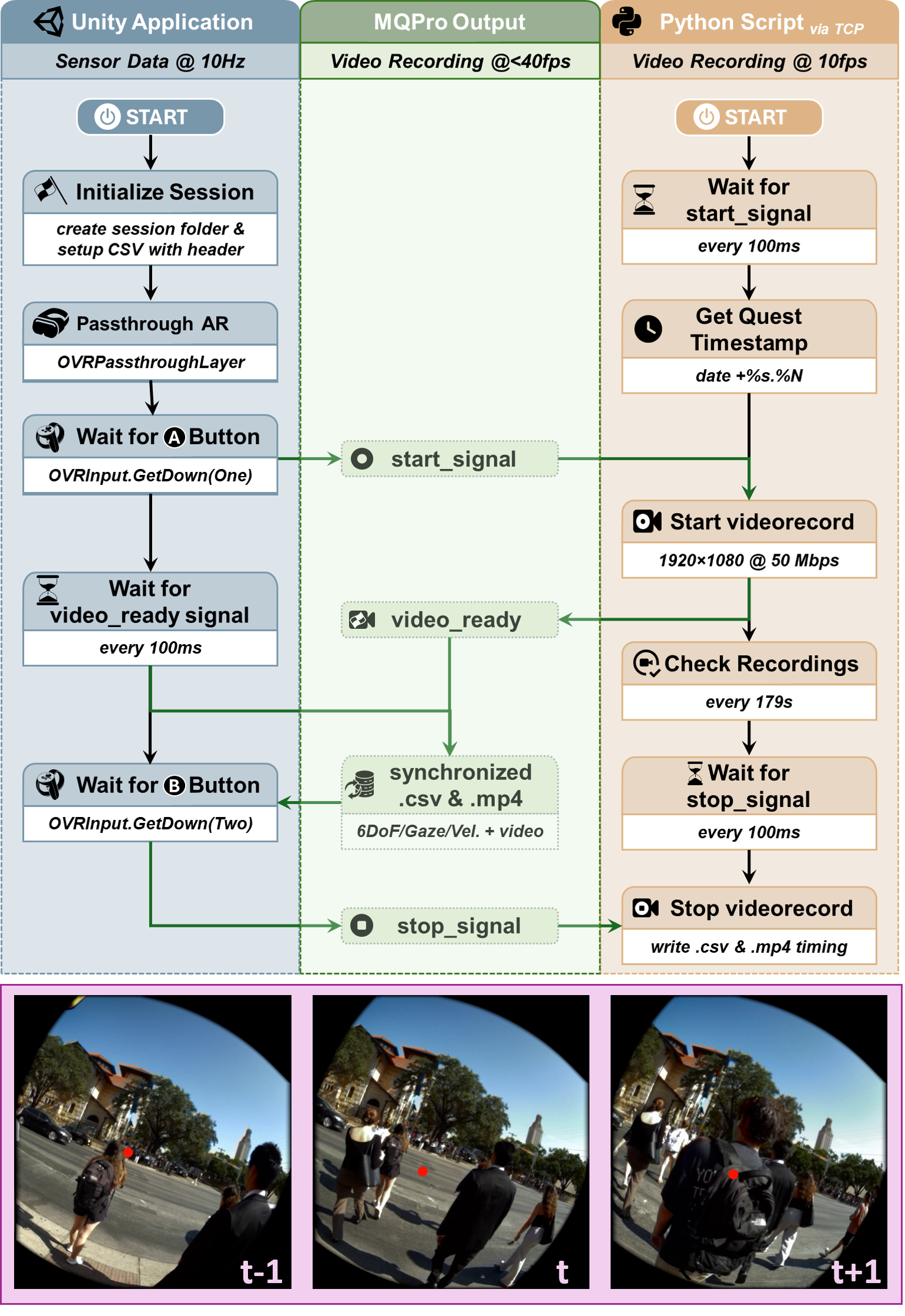}
  \caption{Data collection map for real-time time-synchronized data.}
  \label{Fig3}
\end{figure}

As shown in Figure \ref{Fig3}, all data streams are timestamped to enable accurate temporal alignment. 
Data are collected at 10 Hz, which preserves fine-grained head and eye movements. 
Video frames are matched to the nearest timestamp to pair each image with the corresponding pose, head orientation, and gaze information.
Recordings are segmented into fixed-length sequences containing 2-second observations (20 frames) and 4-second prediction horizons (40 frames). 
To prevent temporal data leakage, all train, validation, and test splits are performed at the session level, with each session assigned exclusively to one single split. 
In total, approximately nine hours of data are recorded to validate the robustness and reliability of the XR-DT data pipeline.

\subsection{Multi-modal Human Motion Prediction} \label{sec:atlas}

We introduce Attention-based Trajectory Learning with Anticipatory Sensing (ATLAS), a Transformer-based model for egocentric trajectory prediction that jointly reasons over four modalities from the MQP headset (Section~\ref{sec:system}).

\subsubsection{Multi-Modal Input Streams}

ATLAS operates on two sensor streams--tracking data (6-DoF pose and eye gaze) and egocentric RGB video--from which four inputs are derived over $T_{\text{obs}} = 20$ frames.
\textbf{(a) Ego-Displacement ($\mathbf{H}$):} Frame-to-frame displacement $\mathbf{h}_t = (\Delta x_t, \Delta y_t, \Delta \theta_t)$ from the headset's 6-DoF pose, encoding speed and turning direction in a location-invariant form.
\textbf{(b) Social Context ($\mathbf{P}$):} 2D body keypoints of nearby pedestrians extracted via ViTPose~\cite{xu2022vitpose}, capturing proximity, facing direction, and movement from the wearer's viewpoint.
\textbf{(c) Scene Context ($\mathbf{C}$):} semantic segmentation (Mask2Former~\cite{cheng2022masked}) compressed through a convolutional autoencoder into a compact feature vector encoding walkable surfaces, obstacles, and road boundaries.
\textbf{(d) Gaze Intent ($\mathbf{G}$):} 2D fixation point $\mathbf{g}_t = (u^g_t, v^g_t)$ in pixel coordinates from the binocular eye tracker. This signal shares the same pixel space as the social and scene streams, enabling the model to learn spatial associations between gaze direction and surrounding context. Gaze anticipates locomotor trajectory by 1--2\,s~\cite{land2006eye}, as a powerful early indicator of turning intent.

\subsubsection{Architecture Overview}

\begin{figure}[t]
  \centering
  \includegraphics[width=\linewidth]{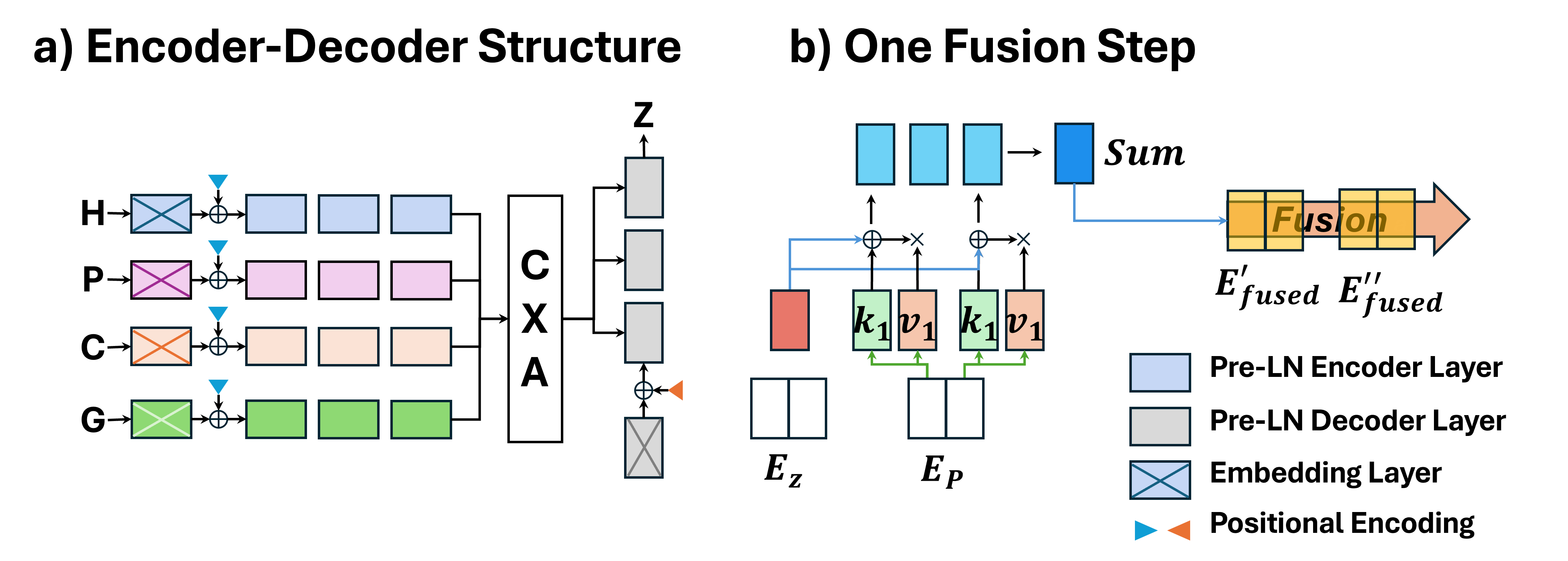}
  \caption{Architecture of multi-modal human motion prediction model.}
  \label{Fig4}
\end{figure}

As shown in Figure~\ref{Fig4}, each modality $\mathbf{m} \in \{\mathbf{H},\mathbf{P},\mathbf{C},\mathbf{G}\}$ is projected to dimension $d=512$ via a modality-specific linear embedding $\psi_m(\cdot)$, combined with sinusoidal positional encoding to preserve temporal ordering, and processed by a dedicated 3-layer Pre-LN Transformer encoder~\cite{xiong2020layer} ($n_h=8$ heads), producing $\mathbf{E}_m \in \mathbb{R}^{T_{\text{obs}} \times d}$. The representations are fused via cascaded cross-attention:
\begin{equation}
    \mathbf{E}_{\text{fused}} = \text{CXA}(\text{CXA}(\mathbf{E}_H,\, \mathbf{E}_P),\, \mathbf{E}_C),
    \label{eq:cxa}
\end{equation}
where $\text{CXA}(\mathbf{A},\mathbf{B})$ denotes cross-attention with $\mathbf{A}$ as queries, $\mathbf{B}$ as keys/values, and a residual connection. The displacement encoding first queries social stream, then scene stream.

\begin{algorithm}[htbp!]
\caption{HA-MPPI Control Pipeline}
\label{alg:ha_mppi}
\begin{algorithmic}[1]
\small
\REQUIRE Current robot state $\boldsymbol{s}_{\text{init}}$, goal state $\boldsymbol{s}_f$, rollouts $K$, samples $N_{\mathrm{mc}}$, horizon $N$, number of pedestrians $N_h$, inverse temperature $\lambda$, safety radius $r$, risk level $\sigma$, static covariance $\boldsymbol{\Sigma}_u$, time-varying covariance $\boldsymbol{\Sigma}_h(t) = \boldsymbol{Cov}(\boldsymbol{e}_t)$, nominal control sequence $\mathbb{U} = \{\boldsymbol{u}_t\}_{t=0}^{N-1}$, human observations $\mathcal{Z}_{t-T_{\mathrm{obs}}+1:t}=\{\mathcal{H}^o, \mathbf{I}^o, \mathbf{g}^o\}_{o=1}^{N_h}$
\ENSURE Optimal control action $\boldsymbol{u}^*_0$ to execute

\STATE \textit{// 1. Multi-Modal Human Motion Prediction}
\FOR{each pedestrian $o \in \{1, \dots, N_h\}$}
    \STATE $\{\hat{\boldsymbol{h}}_t^o\}_{t=1}^N \leftarrow \text{ATLAS}(\mathcal{Z}_{t-T_{\mathrm{obs}}+1:t},o)$ \COMMENT{via Section.~\ref{sec:atlas}}
\ENDFOR

\STATE \textit{// 2. Chance-Constrained MPPI Control}
\FOR{$k = 1$ \TO $K$ \textbf{in parallel}}
    \STATE Initialize state $\boldsymbol{s}_0^k \leftarrow \boldsymbol{s}_{\text{init}}$ and cost $J(\boldsymbol{\tau}_k) \leftarrow 0$
    \STATE Sample control noise $\{\delta\boldsymbol{u}_t^k\}_{t=0}^{N-1} \sim \mathcal{N}(0, \boldsymbol{\Sigma}_u)$
    
    \FOR{$t = 0$ \TO $N-1$}
        \STATE Propagate dynamics $\boldsymbol{s}_{t+1}^k \leftarrow f(\boldsymbol{s}_t^k, \boldsymbol{u}_t + \delta\boldsymbol{u}_t^k)$
        \STATE Accumulate cost $J(\boldsymbol{\tau}_k) \leftarrow J(\boldsymbol{\tau}_k) + J_t(\boldsymbol{s}_t^k, \boldsymbol{u}_t, \delta\boldsymbol{u}_t^k)$
        
        \FOR{each pedestrian $o \in \{1, \dots, N_h\}$}
            \STATE Extract 2D positions $\boldsymbol{p}_t^k$ from $\boldsymbol{s}_t^k$ and $\hat{\boldsymbol{p}}_t^o$ from $\hat{\boldsymbol{h}}_t^o$
            \STATE Sample residuals $\boldsymbol{e}_t^n \sim \mathcal{N}(0, \boldsymbol{\Sigma}_h(t)), n \in \{1, \dots, N_{\mathrm{mc}}\}$
            \STATE Calculate risk via error sampling
            $P_{\text{safety}} \leftarrow\frac{1}{N_{\mathrm{mc}}}\sum_{n=1}^{N_{\mathrm{mc}}} \mathbb{I}\!\left(\left\|\boldsymbol{p}_{t}^{k}-\left(\hat{\boldsymbol{p}}_{t}^{o}+\boldsymbol{e}_{t}^{n}\right)\right\|_2 \ge r\right)$
            \IF{$P_{\text{safety}} < 1 - \sigma$}
                \STATE Add high penalty $J(\boldsymbol{\tau}_k) \leftarrow J(\boldsymbol{\tau}_k) + J_{\text{penalty}}$
            \ENDIF
        \ENDFOR
    \ENDFOR
    \STATE Add terminal cost $J(\boldsymbol{\tau}_k) \leftarrow J(\boldsymbol{\tau}_k) + J_T(\boldsymbol{s}_N^k)$
\ENDFOR

\STATE \textit{// 3. Sampling \& Control Sequence Update}
\FOR{$k = 1$ \TO $K$}
    \STATE Compute trajectory weight $w_k \leftarrow \exp\left(-\frac{1}{\lambda} \left[J(\boldsymbol{\tau}_k) - \min_{k \in K}J(\boldsymbol{\tau}_k)\right]\right)$
\ENDFOR

\FOR{$t = 0$ \TO $N-1$}
    \STATE Update control input $\boldsymbol{u}_t \leftarrow \boldsymbol{u}_t + \frac{\sum_{k=1}^K w_k \delta \boldsymbol{u}_t^k}{\sum_{k=1}^K w_k}$
\ENDFOR
\STATE Conduct Savitzky-Golay filter $\mathbb{U} \leftarrow \text{SGFilter}(\mathbb{U})$

\STATE \textit{// 4. Receding Horizon Shift}
\STATE Apply control input $\boldsymbol{u}^*_0 \leftarrow \boldsymbol{u}_0$
\FOR{$t = 0$ \TO $N-2$}
    \STATE Shift control sequence $\boldsymbol{u}_t \leftarrow \boldsymbol{u}_{t+1}$
\ENDFOR
\STATE Initialize terminal control $\boldsymbol{u}_{N-1} \leftarrow \boldsymbol{u}_{\text{init}}$

\RETURN $\boldsymbol{u}^*_0$
\end{algorithmic}
\vspace{-2pt}
\end{algorithm}

The final fusion incorporates gaze via a novel TGXA mechanism. Standard cross-attention treats all temporal positions equally, ignoring the known offset between gaze and body motion. TGXA augments the attention scores with a learned temporal bias $\mathbf{B} \in \mathbb{R}^{T_{\text{obs}} \times T_{\text{obs}}}$:
\begin{equation}
    \mathbf{E}_{\text{fused}} \leftarrow \mathbf{E}_{\text{fused}} + \text{softmax}\!\left(\frac{\mathbf{Q}\mathbf{K}^\top}{\sqrt{d_k}} + \mathbf{B}\right) \mathbf{V},
    \label{eq:tgxa}
\end{equation}
where $\mathbf{Q} = \mathbf{W}_Q \mathbf{E}_{\text{fused}}$, $\mathbf{K} = \mathbf{W}_K \mathbf{E}_G$, $\mathbf{V} = \mathbf{W}_V \mathbf{E}_G$, and $d_k = d/n_h = 64$. The bias $\mathbf{B}[i,j]$ is initialized with a Gaussian centered at $\Delta \approx 10$ frames, reflecting the observed gaze-motion lag, and learned end-to-end with one matrix per attention head. This allows the model to capture the anticipatory pattern that \textit{gaze at time $t-\Delta$ is the strongest predictor of walking direction at time $t$}.

The fused encoding $\mathbf{E}_{\text{fused}}$ serves as memory for a 3-layer Transformer decoder that autoregressively generates $T_{\text{pred}}=40$ frames of future states. At each step, the decoder attends to $\mathbf{E}_{\text{fused}}$ via cross-attention and produces a hidden state $\mathbf{h}^{\text{dec}}_t \in \mathbb{R}^d$. An output MLP $\gamma(\cdot)$ maps this to the predicted state:
\begin{equation}
    \hat{\mathbf{h}}_t = (\hat{x}_t,\, \hat{y}_t,\, \hat{\theta}_t) = \gamma(\mathbf{h}^{\text{dec}}_t).
    \label{eq:atlas_output}
\end{equation}
The model is trained end-to-end by minimizing:
\begin{equation}
    \mathcal{L} = \frac{1}{T_{\text{pred}}} \sum_{t=1}^{T_{\text{pred}}} \| \hat{\mathbf{h}}_t - \mathbf{h}_t \|_2^2.
    \label{eq:atlas_loss}
\end{equation}
ATLAS runs in under 5\,ms per human on a single GPU, and its predicted trajectory is used directly as $\hat{\boldsymbol{h}}^o_t$ in the HA-MPPI chance constraints (Algorithm~\ref{alg:ha_mppi}, Line~3).

\subsection{Human-Aware Model Predictive Path Integral Algorithm}

Algorithm \ref{alg:ha_mppi} presents the HA-MPPI framework. 
The implementation details of each module have been explained.

\section{Experiments}


\subsection{Human Trajectory Prediction Experiments}

We evaluate the proposed ATLAS on the self-collected dataset described in Section~\ref{subsubsec:data}. 
Prediction quality is assessed using Average Displacement Error (ADE) and Final Displacement Error (FDE) in meters. 
We conduct an ablation study by progressively adding modalities to isolate the contribution of each stream and the TGXA mechanism. 

\begin{figure}[htbp!]
  \centering
  \includegraphics[width=0.7\linewidth]{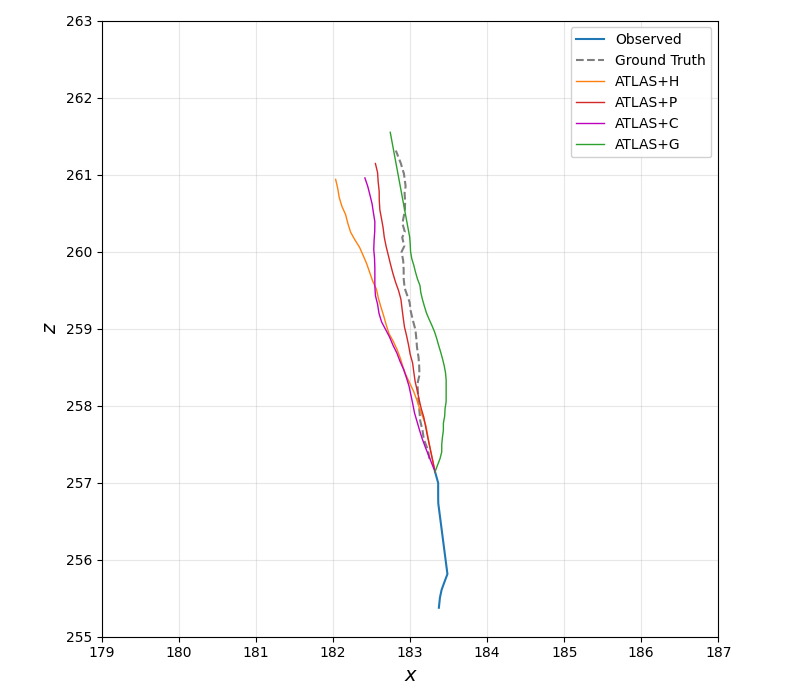}
  \caption{4-second human trajectory prediction.}
  \label{Fig5}
\end{figure}

\begin{table}[htbp!]
\centering
\caption{Ablation study.}
\label{tab:ablation}
\begin{tabular}{lccccc}
\hline
\textbf{Configuration} & $\mathbf{H}$ & $\mathbf{P}$ & $\mathbf{C}$ & $\mathbf{G}$ & \textbf{ADE / FDE (m)}$\downarrow$ \\
\hline
Displacement only          & \checkmark &            &            &            & 0.66 / 1.18 \\
+ Social (CXA)             & \checkmark & \checkmark &            &            & 0.59 / 1.14 \\
+ Scene (CXA)              & \checkmark & \checkmark & \checkmark &            & 0.56 / 0.97 \\
+ Gaze (CXA)        & \checkmark & \checkmark & \checkmark & CXA        & 0.48 / 0.90 \\
+ Gaze (TGXA)              & \checkmark & \checkmark & \checkmark & \checkmark & \textbf{0.44 / 0.86} \\
\hline
\end{tabular}
\end{table}

Figure~\ref{Fig5} visualizes trajectory predictions across ablation configurations. 
The displacement-only model drifts increasingly from the ground truth over the 4-second horizon, particularly at turning points where future direction cannot be inferred from past displacement alone. 
Adding social and scene context reduces this drift by constraining predictions to plausible paths. 
The most visible improvement occurs when gaze is incorporated: the full ATLAS model with TGXA maintains close alignment with the actual trajectory through turns, demonstrating that anticipatory gaze modeling produces sharper and more accurate turn predictions.

\begin{table*}[ht]
    \centering
        \caption{Performance comparison of different motion planning methods under different pedestrian densities.}
    \medskip
    \resizebox{\linewidth}{!}{
    \begin{tabular}{clccccc}
        \hline
        \textbf{\# of Pedestrians} & \textbf{Method} & \textbf{Robot Duration (s)$\downarrow$} & \textbf{Robot Speed (m/s)$\uparrow$} & \textbf{Human Duration (s)$\downarrow$} & \textbf{Human Speed (m/s)$\uparrow$} & \textbf{Min Distance (m)$\uparrow$}  \\
        \hline
        \multirow{5}{*}{1} & SH-MPC \cite{de2025scenario} & 20.4 $\pm$ 1.1 & 0.74 $\pm$ 0.08 & 12.3 $\pm$ 0.9 & 1.32 $\pm$ 0.10 & 0.61 $\pm$ 0.05 \\
        & Vanilla MPPI \cite{williams2017model} & \textbf{18.9 $\pm$ 0.6} & \textbf{0.83 $\pm$ 0.05} & 13.3 $\pm$ 1.0 &  1.25 $\pm$ 0.12 & 0.48 $\pm$ 0.05 \\
        & DRA-MPPI \cite{trevisan2025dynamic}  & 20.0 $\pm$ 0.9 & 0.76 $\pm$ 0.07 & 12.1 $\pm$ 0.8 &  1.34 $\pm$ 0.09 & \underline{0.64 $\pm$ 0.05} \\
        & \textbf{HA-MPPI w/o XR-DT (Ours)} & 19.9 $\pm$ 0.8 & 0.78 $\pm$ 0.07 & \underline{11.4 $\pm$ 0.7} & \underline{1.40 $\pm$ 0.08} & \underline{0.64 $\pm$ 0.06} \\
        & \textbf{HA-MPPI w/ XR-DT (Ours)} & \underline{19.2 $\pm$ 0.7} & \underline{0.81 $\pm$ 0.06} & \textbf{10.7 $\pm$ 0.6} & \textbf{1.46 $\pm$ 0.07} & \textbf{0.75 $\pm$ 0.05} \\
        \hline
        \multirow{5}{*}{2} & SH-MPC \cite{de2025scenario} & 21.1 $\pm$ 1.3 & 0.72 $\pm$ 0.09 & 13.2 $\pm$ 1.1 & 1.26 $\pm$ 0.11 & 0.58 $\pm$ 0.05\\
        & Vanilla MPPI \cite{williams2017model} & \textbf{19.4 $\pm$ 0.8} & \textbf{0.81 $\pm$ 0.06} & 14.1 $\pm$ 1.2 & 1.20 $\pm$ 0.13 & 0.44 $\pm$ 0.05 \\
        & DRA-MPPI \cite{trevisan2025dynamic} & 20.8 $\pm$ 1.1 & 0.73 $\pm$ 0.08 & 12.6 $\pm$ 0.9 & 1.30 $\pm$ 0.10 & 0.60 $\pm$ 0.06 \\
        & \textbf{HA-MPPI w/o XR-DT (Ours)} & 20.6 $\pm$ 1.0 & 0.76 $\pm$ 0.08 & \underline{11.7 $\pm$ 0.8}  & \underline{1.38 $\pm$ 0.09} & \underline{0.63 $\pm$ 0.06} \\
        & \textbf{HA-MPPI w/ XR-DT (Ours)} & \underline{20.0 $\pm$ 0.9} & \underline{0.78 $\pm$ 0.07} & \textbf{11.1 $\pm$ 0.7} & \textbf{1.43 $\pm$ 0.08} & \textbf{0.74 $\pm$ 0.06} \\
        \hline
    \end{tabular}}
    \label{tab:comparison}
\end{table*}

\begin{figure*}[htbp!]
  \centering
  \includegraphics[width=0.75\linewidth]{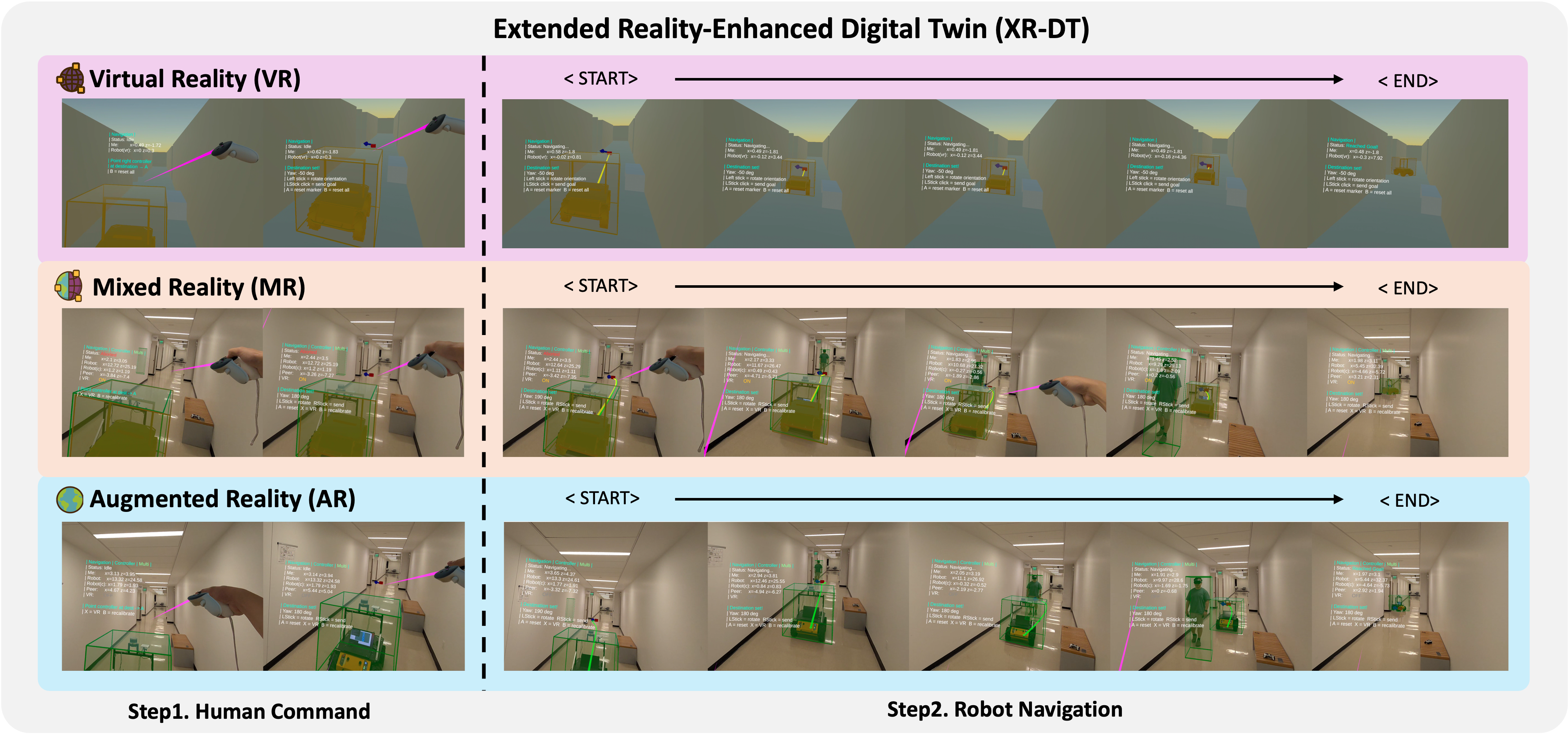}
  \caption{Demonstration of the XR-DT framework in a real-world robot navigation scenario.}
  \label{Fig6}
\end{figure*}

As shown in Table~\ref{tab:ablation}, adding social context $\mathbf{P}$ reduces ADE by 10.6\% and FDE by 3.4\%, indicating that the awareness of nearby pedestrians helps the model anticipate yielding and avoidance behavior along the trajectory.
Scene context $\mathbf{C}$ provides a notable decrease in FDE, confirming that scene information remains relevant over the full prediction horizon.
Gaze $\mathbf{G}$ produces the largest single-modality gain, reducing ADE from 0.56 to 0.48\,m (14.3\%) and FDE from 0.97 to 0.90\,m (7.2\%), confirming that gaze is the most informative contextual signal for egocentric trajectory prediction. 
TGXA further reduces ADE from 0.48 to 0.44\,m (8.3\%), while FDE from 0.90 to 0.86\,m (4.4\%), showing that TGXA learns to attend to gaze fixations that occurred approximately 1-second before the corresponding displacement, producing more accurate predictions along the entire trajectory. 
Overall, the full ATLAS model achieves a 33.3\% ADE reduction and 27.1\% FDE reduction over the displacement-only baseline.

\subsection{Robot Motion Planning Experiments}

Real-world experiments are conducted in a narrow corridor where pedestrians walk alongside the robot. 
30 trials are performed under each of the two conditions.
We compare our HA-MPPI against three baselines: (1) Vanilla MPPI  \cite{williams2017model}, (2) Safe Horizon MPC (SH-MPC) \cite{de2025scenario}, and (3) Dynamic Risk-Aware MPPI (DRA-MPPI) \cite{trevisan2025dynamic}.
Following common practice in \cite{trevisan2025dynamic}, the baselines' pedestrian positions are estimated using a Kalman filter and assume constant velocity predictions for future motion.
The following four metrics are considered for evaluation: (1) Robot Duration (s), (2) Robot Speed (m/s), (3) Human Duration (s), (4) Human Speed (m/s)， and (5) Min Distance (m) (between human and robot). The quantitative results for scenarios with one and two pedestrians are summarized in Table \ref{tab:comparison}.

Quantitatively, Vanilla MPPI achieves the shortest robot travel times across both pedestrian densities, but this is accompanied by more aggressive behavior and less consideration of human motion, as reflected by longer human travel times and lower average human speeds. 
SH-MPC and DRA-MPPI exhibit more conservative behavior, resulting in longer robot durations and reduced average speeds, but with shorter path. 
In contrast, our HA-MPPI provides a balanced trade-off between efficiency and safety, achieving competitive robot performance while maintaining more cautious interactions with improved human-centric performance. 
Importantly, integrating the XR-DT interface further amplifies this effect, achieving the shortest human duration and highest human speed in both density settings, suggesting that exposing the robot's intent and future plans allows pedestrians to adapt their motion more efficiently and confidently.
Notably, across all trials, no collisions are recorded.

\begin{table}[t]
\centering
\vspace{6pt}
\caption{User study.}
\label{tab:user}
\begin{tabular}{lccc}
\hline
\textbf{Method} & \textbf{Interpretability}$\uparrow$ & \textbf{Trust}$\uparrow$ & \textbf{Safety}$\uparrow$ \\
\hline
HA-MPPI w/o XR-DT          & 2.41     &     2.20     & 2.87 \\
HA-MPPI w/ XR-DT           & \textbf{4.51}  &     \textbf{4.75}     &  \textbf{3.54} \\
\hline
\end{tabular}
\end{table}

\subsection{User Study}

Figure \ref{Fig6} illustrates a representative interaction within the XR-DT framework: (1) Green/Yellow line (planned trajectory in real/virtual world) shows that after the human sends a command to the robot, the proposed HA-MPPI generates smooth and anticipatory motions while respecting corridor boundaries. (2) Green/Yellow bounding box (cues for real/virtual robot) confirms that the XR-DT system correctly aligns the virtual robot's status with the real robot in the physical world, enabling the planner to reason over both the real and virtual representations of the environment.
The XR headset renders the XR-DT overlays at 30 fps, remaining stable without noticeable jitter or delay.

We distribute 60 questionnaires about the experiences with XR-DT and ultimately collect 53 valid responses.
The user study is conducted across three dimensions: Interpretability, Trust, and Safety. 
Participants are asked to rate on a scale of 1 to 5. 
The results are presented in Table \ref{tab:user}, which shows that our method has received human approval for its comprehensive capability of interpretable, trustworthy, and adaptive HRI.
Unlike traditional ``black-box'' planners, the XR-DT interface externalizes the robot's intent. 
This visual feedback shifts HRI from reactive avoidance to predictive collaboration. 
By revealing these hidden states, the XR-DT framework mitigates the ``surprise factor'', allowing humans to anticipate and prevent unsafe maneuvers before physical execution, further fostering human trust in the robot.

\section{Conclusion and Future Work}

This paper introduces an XR-DT framework for mobile robots, which unifies AR-, VR-, and MR-based perception-interaction loops to enable safe, interpretable, and adaptive HRI. 
By integrating real-world sensing with simulation-driven predictive intelligence, the framework establishes bi-directional understanding between humans and robots. 
In addition, this paper introduces a novel HA-MPPI control model, incorporating a chance-constraint MPPI with a multi-modal human motion prediction model (ATLAS).
Experimental evaluations demonstrate accurate human trajectory prediction and safe and efficient robot motion planning, validating the feasibility of the HA-MPPI control model within the XR-DT framework. 
Case study showcases that our XR-DT receives human approval for improved interpretable, trustworthy, and safe HRI.
Future work will focus on extending the platform for multi-human and multi-robot settings, improving generalization in open-ended environments.










\bibliographystyle{IEEEtran}
\bibliography{IEEEexample}

\end{document}